\documentclass[a4paper,twoside]{article}

\usepackage{epsfig}
\usepackage{subcaption}
\usepackage{calc}
\usepackage{amssymb}
\usepackage{amstext}
\usepackage{amsmath}
\usepackage{amsthm}
\usepackage{multicol}
\usepackage{pslatex}
\usepackage{apalike}
\usepackage{SCITEPRESS}     

\usepackage{cite}
\usepackage{hyperref}

\begin{document}

\title{Evaluation of the Synthetic Electronic Health Records}

\author{\authorname{Emily Muller\sup{1}, Xu Zheng\sup{2} and Jer Hayes\sup{2}}
\affiliation{\sup{1}Department of Epidemiology and Biostatistics, Imperial College London, UK}
\affiliation{\sup{2}Accenture Labs, Dublin, Ireland}
\email{emuller@ic.ac.uk, \{xu.b.zheng, jeremiah.hayes\}@accenture.com}
}


\abstract{Generative models have been found effective for data synthesis due to their ability to capture complex underlying data distributions. The quality of generated data from these models is commonly evaluated by visual inspection for image datasets or downstream analytical tasks for tabular datasets. These evaluation methods neither measure the implicit data distribution nor consider the data privacy issues, and it remains an open question of how to compare and rank different generative models. Medical data can be sensitive, so it is of great importance to draw privacy concerns of patients while maintaining the data utility of the synthetic dataset. Beyond the utility evaluation, this work outlines two metrics called \textit{Similarity} and \textit{Uniqueness} for sample-wise assessment of synthetic datasets. We demonstrate the proposed notions with several state-of-the-art generative models to synthesise Cystic Fibrosis (CF) patients' electronic health records (EHRs), observing that the proposed metrics are suitable for synthetic data evaluation and generative model comparison.}

\onecolumn \maketitle \normalsize \setcounter{footnote}{0} \vfill

\section{\uppercase{Introduction}} \label{intro}

\noindent Data is constantly generated in the field of medicine from sources such as biosensors, physiological measurements, genome sequencing and electronic health records (EHRs). Despite this, the data on specific sub-populations may be pretty scarce when aggregating data is expensive if the data is proprietary; or rendered inaccessible due to interoperability standards in the sharing of health data; or illegal to share it has the potential to violate privacy. To overcome these issues, synthetic data are increasingly being used in the healthcare setting \cite{chen2021synthetic}. A desirable data synthesis algorithm in the medical domain should be able to generate new samples that preserve the original data distribution while adhering to properties of privacy. Preserving the data distribution maintains good utility of the synthetic data in downstream analytical tasks while maintaining the privacy level ensures that the synthetic data do not leak information of any single individual.

Deep generative models, such as generative adversarial networks (GANs) and variational autoencoders (VAEs) \cite{xu2018synthesizing, choi2017generating, xu2019modeling, xie2018differentially, yoon2020anonymization}, have been found effective for synthesising medical datasets. These models implicitly parameterise the multivariate distribution of the original data using deep neural networks. In \cite{fiore2019using} a GAN is used to synthesise the underrepresented class of fraudulent credit card cases, observing a maximum increase in classification sensitivity of $3.4$ percentage points when augmenting the small class 3 times its size ($0.55\%$ of the training set). While a conditional GAN can achieve optimal performance on a set of 22 tabular datasets \cite{douzas2018effective}. In an application of augmenting classes of thermal comfort, the authors find two experiments where synthetic data alone has a higher F1 score than the original training data. Similar to \cite{fiore2019using}, the author synthesises the under-represented class. Other work has shown the efficacy of generative networks above traditional methods, such as the Synthetic Minority Over-sampling Technique (SMOTE) and its variants \cite{liu2019wasserstein, ngwenduna2021alleviating, engelmann2020conditional}.

Although the high capacity of generative models makes them good candidates for capturing complex non-linear distributions in the data, their intractable likelihood functions make evaluation difficult. In this work, we propose to use \textit{Similarity} and \textit{Uniqueness} for sample-wise evaluation of Cystic Fibrosis EHRs synthesising. We found increased predictive performance during the experiments when augmenting given datasets with synthetic data, which is further faithfully evaluated by the proposed metrics.

\section{\uppercase{Method}}\label{method}
\noindent We propose to compare generative models from multiple perspectives, including \textit{Uniqueness}, \textit{Similarity} and \textit{Utility}. In this section, we detail the feature extraction procedure of the data, definitions of these evaluation metrics and generative model selection.

\subsection{Feature extraction}
Cystic Fibrosis (CF) is a rare disease that gives rise to different forms of lung dysfunction, eventually leading to progressive respiratory failure. It is a complex disease, and the types and severity of symptoms can differ widely from person to person. In our work, we extract CF patients from the IBM Explorys database with a total of $10074$ patients extracted, representing about 1/3 ($31199$) of all CF patients in the US \cite{USCF}. Patients belong to two subgroups: having died or having received a lung transplant, labeled by value 0; or having survived, labeled by value 1. We remove all samples with no diagnosis codes and duplicates to enhance synthetic diversity. Our final dataset has $3184$ patients, with $\sim 80\%$ belonging to the survived subgroup. The EHRs of these patients are then aggregated over time. For each patient, we assign value 1 to the features that have appeared in the medical history, and value 0 to these features that have never appeared. The medical data is finally represented as a binary matrix where each row corresponds to a patient and each column to a medical feature. The predictive survival outcome is based on these medical features including comorbidity, lung infection, and therapy variables.

\subsection{Uniqueness}
Privacy assurances are essential to prevent the leakage of personal information. However, a synthetic data generator can achieve perfect evaluation scores by simply copying the original training data, thus breaking the privacy guarantee. Differential privacy (DP) is one well-known and commonly researched assurance \cite{dwork2014algorithmic}. DP algorithms limit how much; the output can differ based on whether the input is included; one can learn about a person because their data was included; and confidence about whether someone's data was included. In practice, there are various distance-based metrics to guarantee such assurance. In \cite{alaa2021faithful}, the authors quantify a generated sample as authentic if its closest real sample is at least closer to any other real sample than the generated sample. Extending this to the case of binary variables, we could consider hamming distance. However, since our data is de-personalised and non-identifiable, we assess our generators based on how many exact copies are made. We consider the requirement of privacy as \textit{Uniqueness}: to not simply copy the input data. We calculate the \textit{Uniqueness} of each model by generating a large finite number of samples and reporting the percentage of overlap with the original training data.

\subsection{Similarity}
It is hard to measure the \textit{Similarity} between the synthetic and original datasets with one score because of the multiple features and data types within the data. In our work, the \textit{Similarity} is measured with four sub-metrics, precision, recall, density and coverage. 

In \cite{sajjadi2018assessing} the use of precision and recall metrics to measure the output from generative models is proposed. Precision measures the fidelity - the degree to which generated samples resemble the original data. Furthermore, recall measures diversity - whether generated samples cover the full variability of the original data. The latter is particularly useful for evaluating generative models prone to mode collapse. Precision is defined as the proportion of the synthetic probability distribution that can be generated from the original probability distribution, thus measuring fidelity, and recall symmetrically defines diversity. Precision and recall are effectively calculated as the proportion of samples that lie in support of the comparative distribution, which assumes uniform density across the data. Therefore, alternative metrics have been proposed, such as density and coverage, to ameliorate this issue \cite{kynkaanniemi2019improved, naeem2020reliable, alaa2021faithful}. We employ the definition and implementation of density and coverage metrics from \cite{naeem2020reliable} for a more accurate \textit{Similarity} evaluation. Density and coverage address the lack of robustness to outliers, failure to detect matching distributions and inability to diagnose different types of distribution failure.
\begin{figure*}
     \begin{subfigure}[b]{0.32\textwidth}
         \centering
         \includegraphics[width=\textwidth]{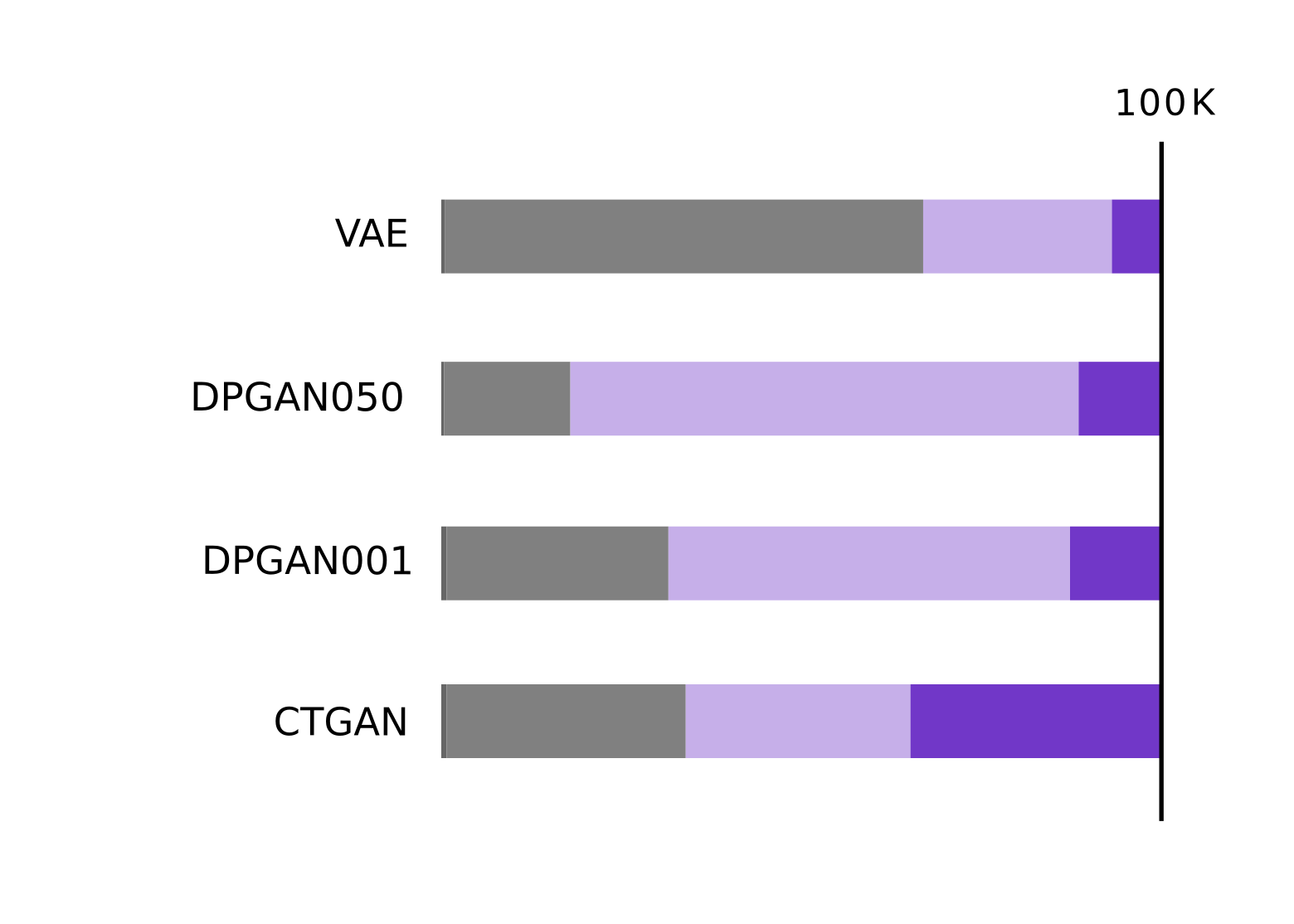}
         \caption{Authenticity proportion}
         \label{fig1:privacy}
     \end{subfigure}
     \hfill
     \begin{subfigure}[b]{0.32\textwidth}
         \centering
         \includegraphics[width=\textwidth]{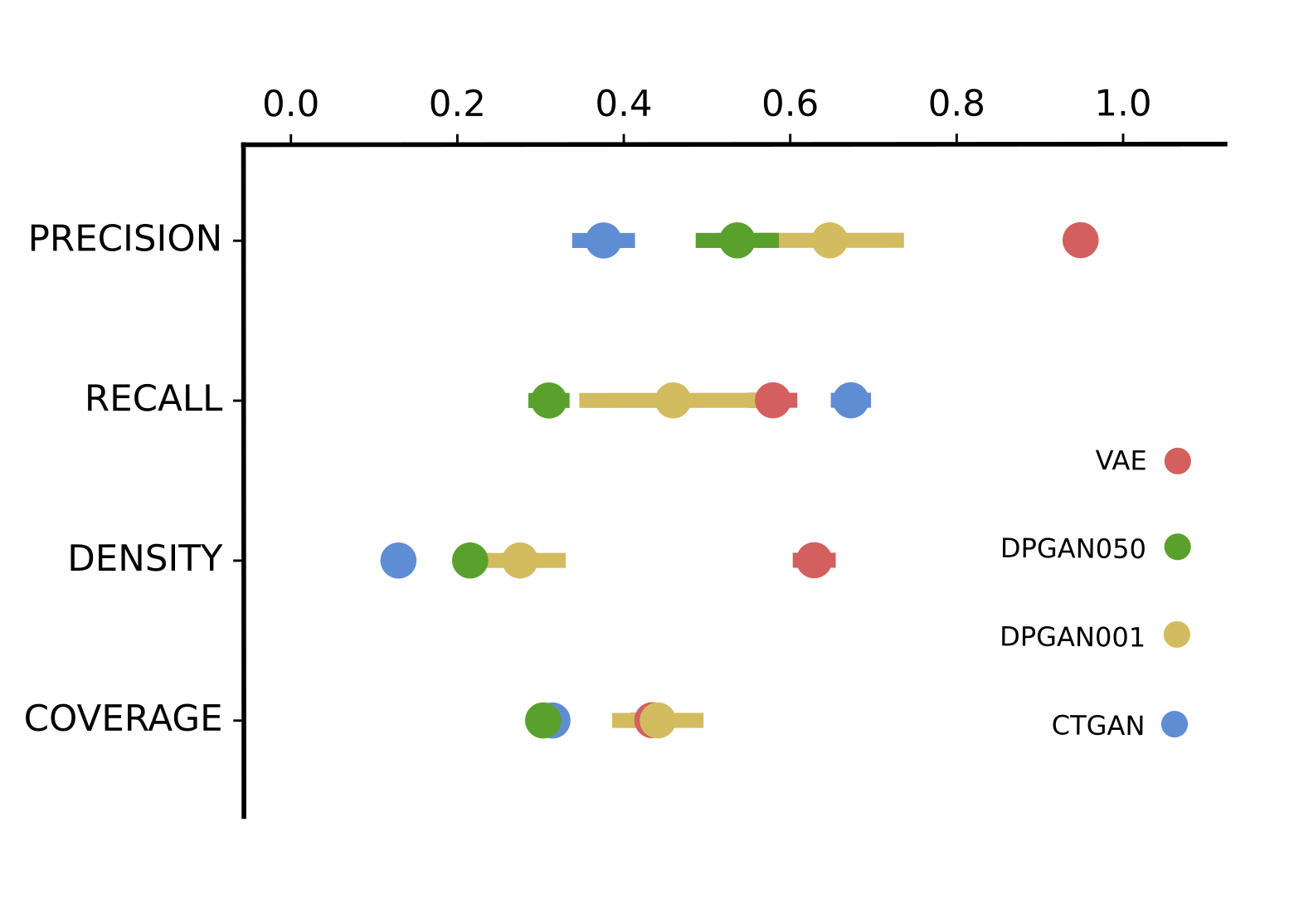} 
         \caption{Similarity metrics}
         \label{fig2:similarity}
     \end{subfigure}
     \begin{subfigure}[b]{0.32\textwidth}
         \centering
         \includegraphics[width=\textwidth]{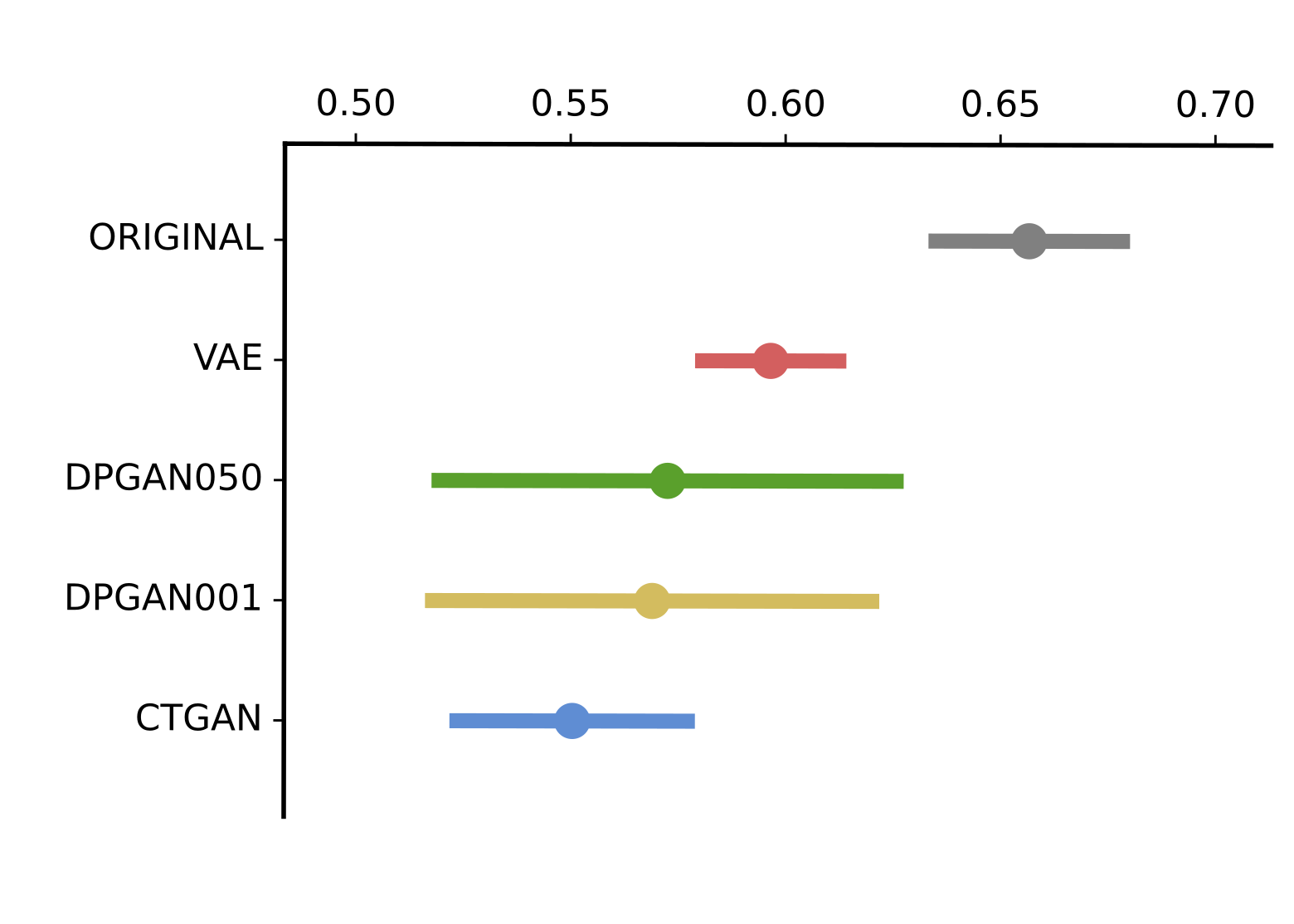}
         \caption{Synthetic data AUC-ROC}
         \label{fig3:synth_auc}
     \end{subfigure}
    \caption{(A) Authenticity of $100k$ samples from each generator. Grey is the proportion of samples that appear in the original training data. Light purple is samples that do not appear in the original training data, and darker purple represents those that are unique. (B) Similarity metrics for each model. For each fold, a dataset matching the size of the original fold with the equivalent proportion of classes is sampled from the unique synthetic dataset (dark blue only). This is repeated 10 times, and similarity metrics show mean and standard deviation over folds and repetitions.}
    \label{fig:stand_alone}
\end{figure*}
\subsection{Utility}
To empirically validate the \textit{Utility} of the generated dataset, we introduce two different training testing settings. \textbf{Setting A:} train the predictive models on the synthetic training set and test the performance of the models on the testing set. \textbf{Setting B:} train on the synthetically augmented balanced-class training set as the original dataset is imbalanced, then test on the testing set. We perform a 5-fold cross-validation, sampling each fold with a proportional representation of each class. After training the synthesisers (described below), we obtain unique synthetic datasets for each fold and model. Setting A was implemented as follows: for each fold, we sampled a synthetic dataset the size of the original dataset with equivalent class proportions to calculate \textit{Utility}. We repeated this ten times resulting in 50 measurements for each model for which we report the mean and standard deviation. For each synthetic dataset, we trained the following classification models: Support Vector Machine, Linear Regression, Naive Bayes, K-Nearest Neighbours and a Random Forest. We chose the classifier which maximised the area under the receiver operating curve (AUC-ROC) for the survived subgroup with label 1 for the holdout test set and compared this to the training with original data. As well as a testing utility on the synthetic dataset alone, we also augmented the original dataset to obtain class balance (Setting B). We performed the same utility experiment outlined above, reporting both areas under the receiver operating curve (AUC-ROC) and accuracy with a balanced class augmented training set.

\subsection{Synthetic Data Generators}
We implemented several SOTA generative models, including variational autoencoders (VAE), Conditional GAN (CTGAN) \cite{xu2019modeling} and Differentially Private GAN (DPGAN) \cite{xie2018differentially} with two privacy levels. DPGAN001 has sigma 0.01 and gradient clip 0.1, while DPGAN050 has sigma 0.5 and gradient clip 0.05. We mainly pick these two extreme privacy settings for the DPGAN model to explore how noise levels impact synthetic datasets. For these models, we attempted to preserve their original architecture as published and adjusted the hyperparameters using grid search, with the optimisation objective to maximise the sum of \textit{Similarity} metrics, including precision, recall, density and coverage. VAE is trained with two hidden layers in the encoder and decoder and ELBO loss minimisation. DPGAN is trained as a typical minimax game between the discriminator and generator with noise injected into the gradient during training with the cross-entropy derived loss. CTGAN addresses sparsity and imbalanced categorical columns by sampling vectors during training and introducing a condition for which the generator learns the conditional distribution.

\section{\uppercase{Results}}\label{results}
\noindent After hyperparameters tuning and model optimisation for each of the 5 separate training folds, we generated 100K synthetic samples from each model. We post-processed each synthetic dataset to obtain only unique samples. For each fold, we sampled a synthetic dataset the size of the original dataset with equivalent class proportions to calculate \textit{Similarity}. We repeated this 10 times resulting in 50 measurements for each metric and model for which we report the mean and standard deviation in the plots.

\begin{figure*}
     \begin{subfigure}[b]{0.49\textwidth}
         \raggedleft
         \includegraphics[width=0.9\textwidth,trim=4 4 4 4,clip]{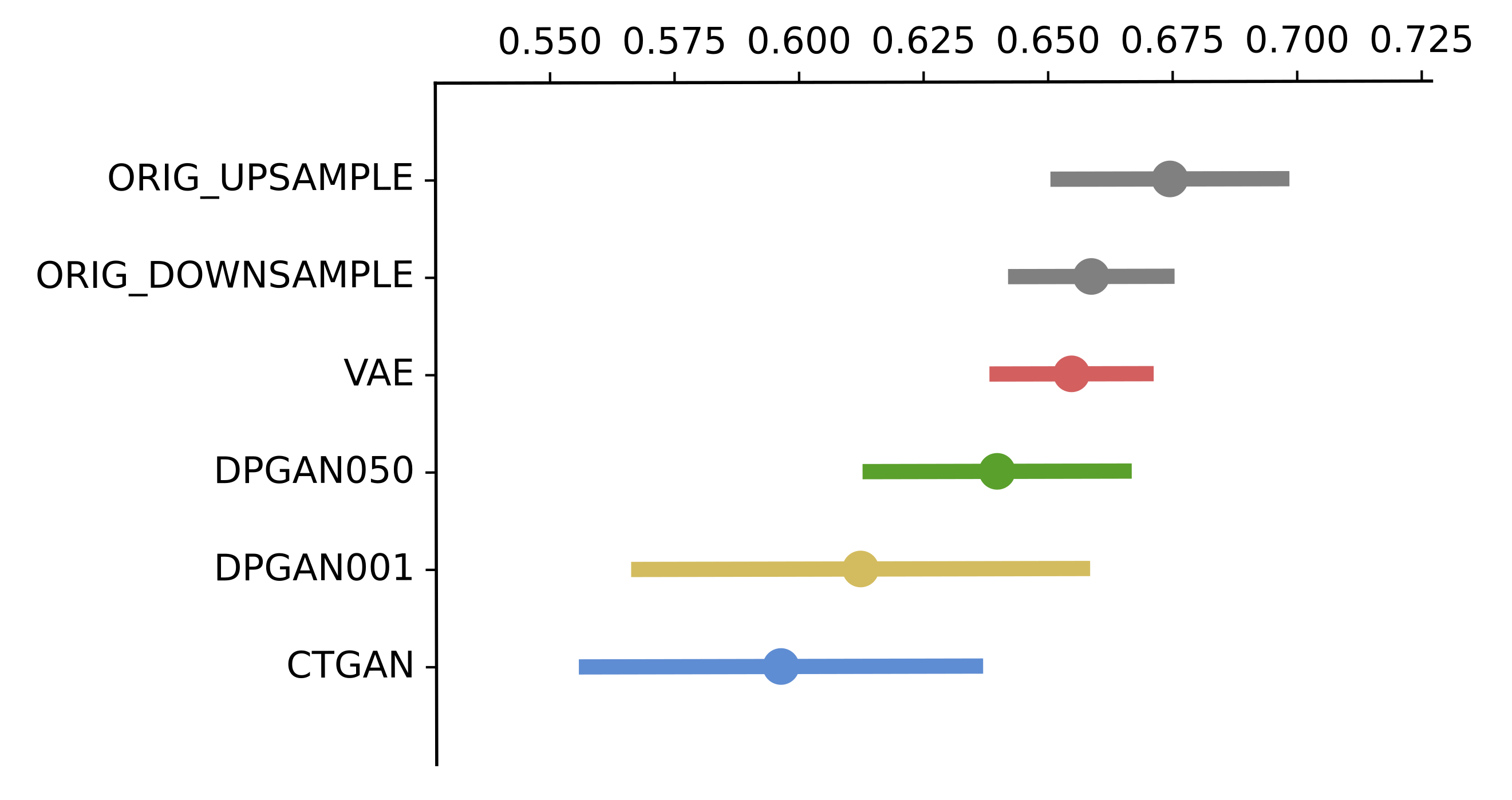}
         \centering
         \caption{AUC-ROC}
         \label{fig:augauc}
     \end{subfigure}
     \hfill
     \begin{subfigure}[b]{0.49\textwidth}
         \raggedright
         \includegraphics[width=0.9\textwidth,trim=4 4 4 4,clip]{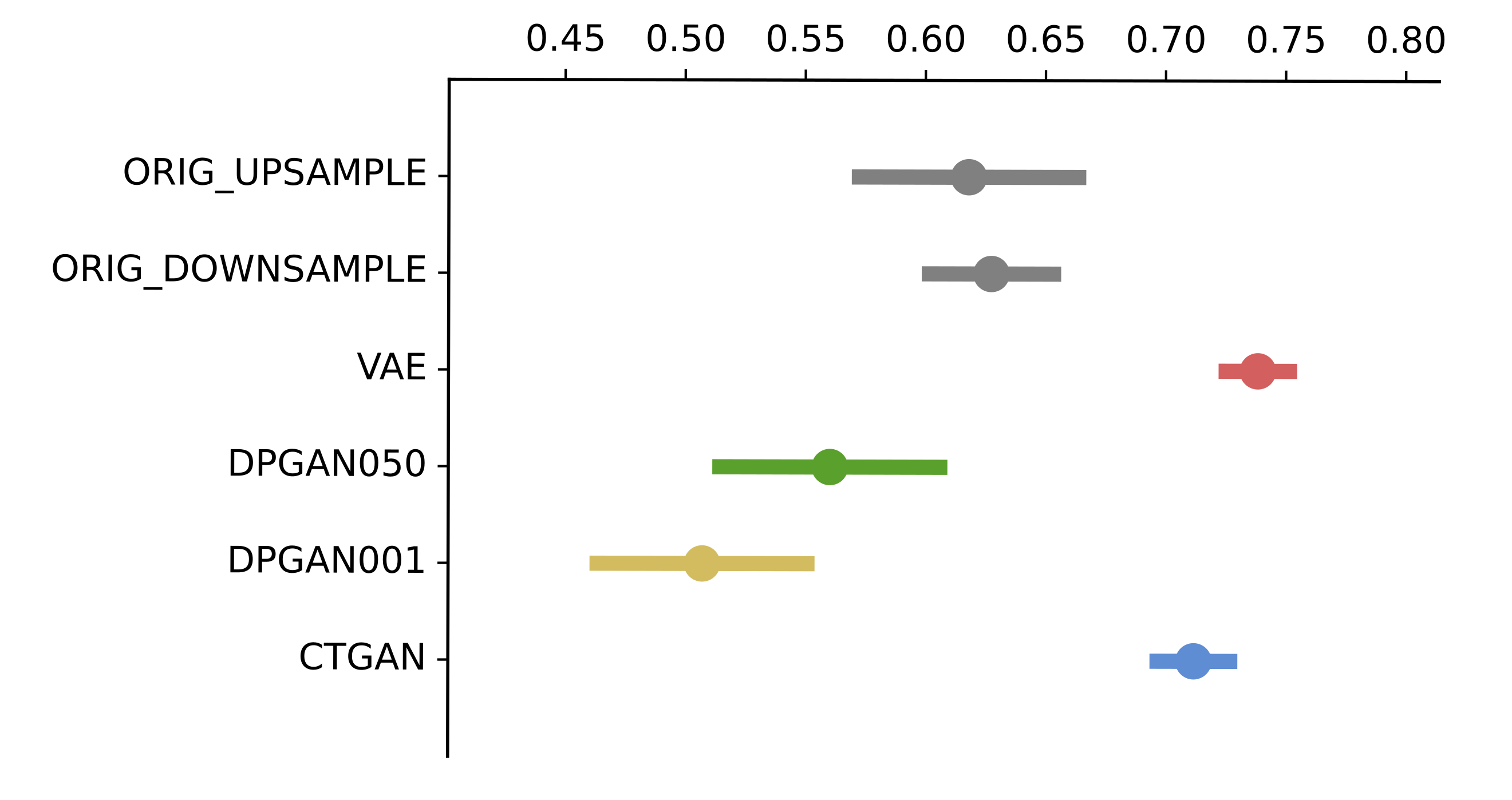}
         \centering
         \caption{Accuracy}
         \label{fig:augacc}
     \end{subfigure}
     \hfill
     \hfill
     \vspace{0.5cm}
     \begin{subfigure}[b]{0.49\textwidth}
         \raggedleft
         \includegraphics[width=0.9\textwidth,trim=4 4 4 4,clip]{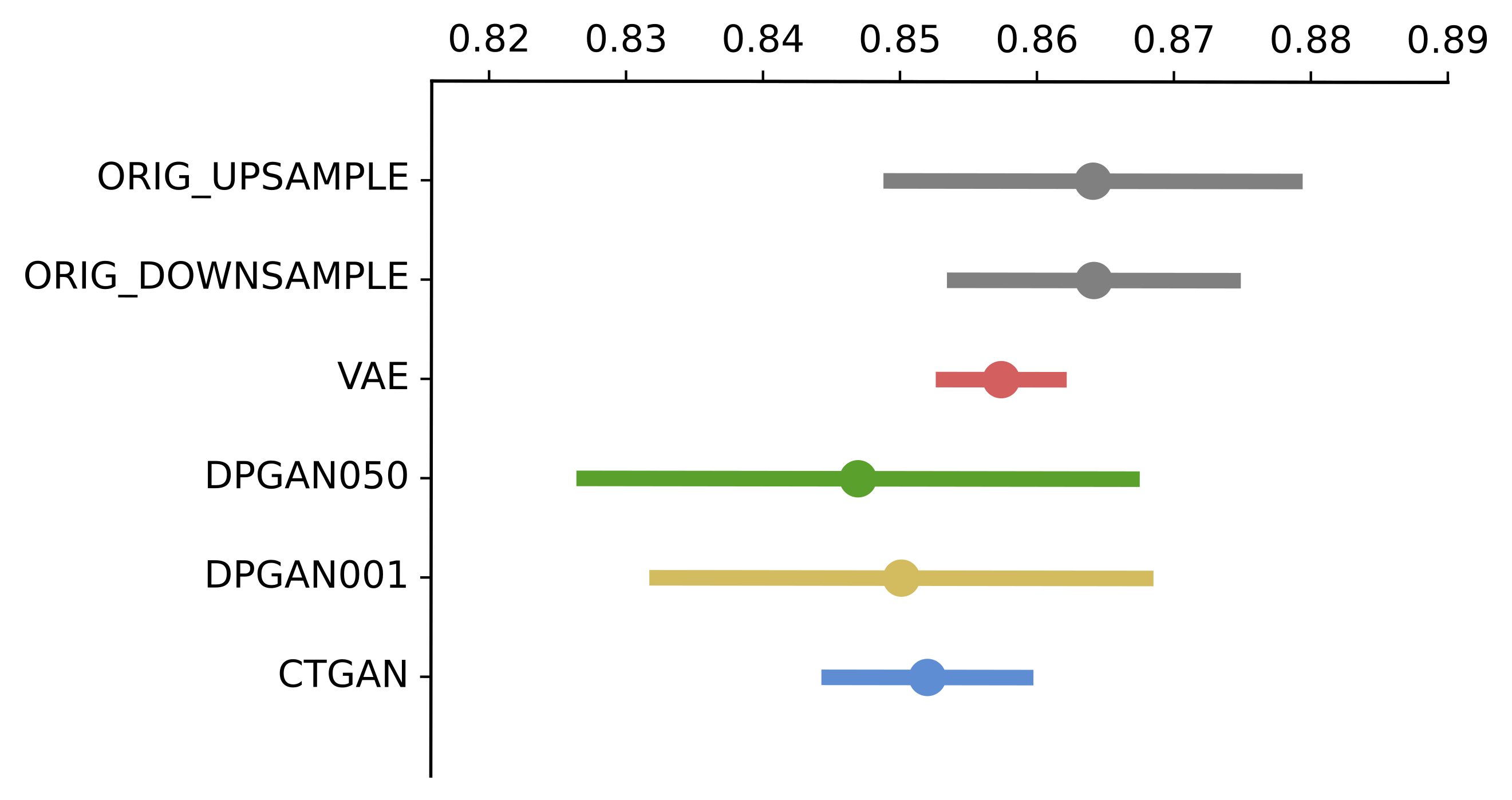}
         \centering
         \caption{Precision}
         \label{fig:aucprec}
     \end{subfigure}
     \begin{subfigure}[b]{0.49\textwidth}
         \raggedright
         \includegraphics[width=0.9\textwidth,trim=4 4 4 4,clip]{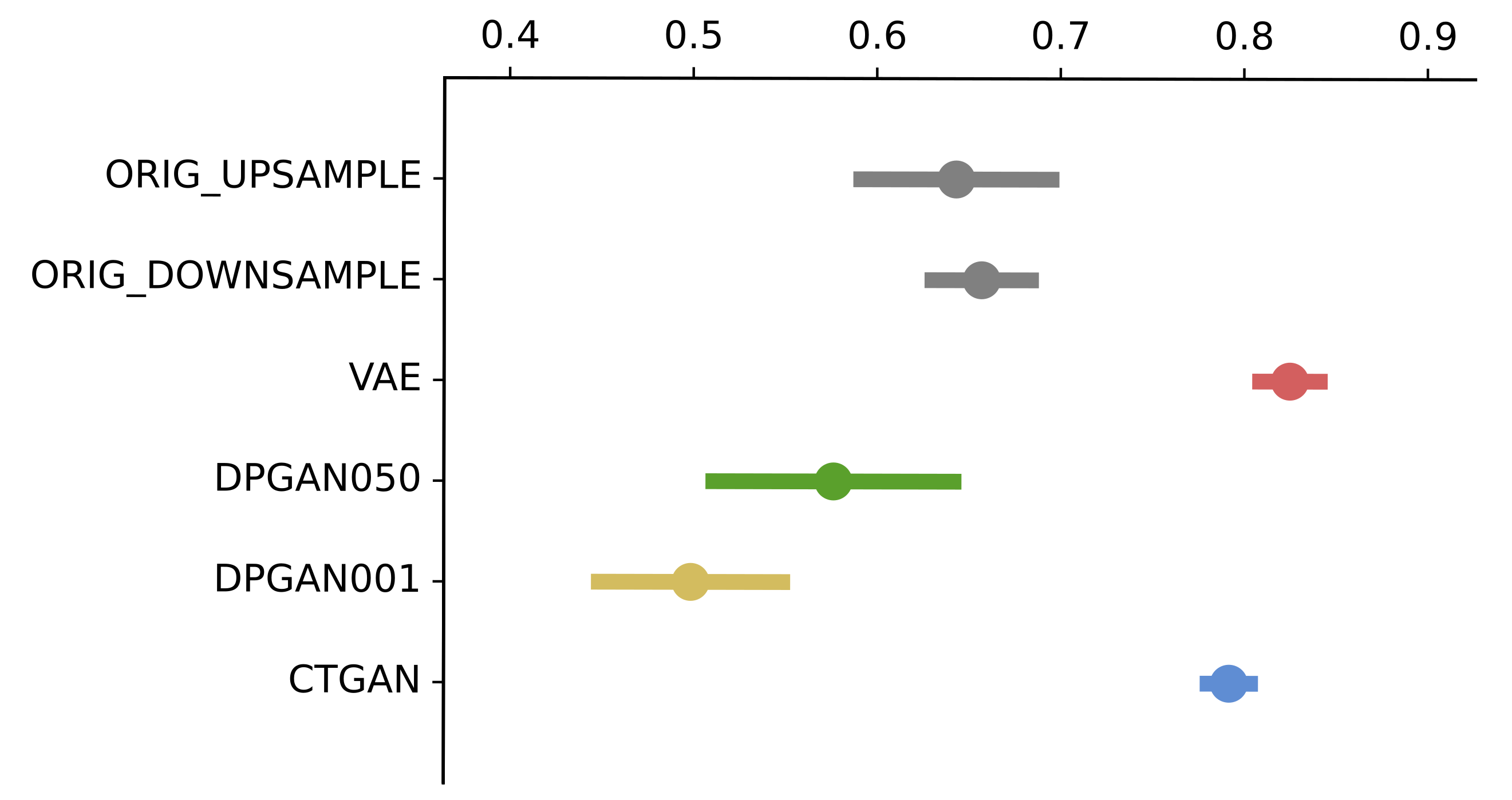}
         \centering
         \caption{Recall}
         \label{fig:augrec}
     \end{subfigure}
    \caption{Performance of original data augmented with synthetic samples. For each fold, synthetic samples are randomly drawn to augment the small class. Where there are not enough of the minor class, the large class is randomly downsampled achieving class balance. Gaussian Naive Bayes, Random Forest, Neural Network, SVM and Logistic Regression classifiers are each trained to predict the hold out test set outcome and results which maximise performance metrics are reported, averaged over each fold and repeated 10 times. The average number of samples for each model and class is $2128$, $419$, $1029$, $1908$, $1974$ and $2128$ for Upsampling, Downsampling, VAE, DPGAN050, DPGAN001, CTGAN respectively.}
    \label{fig:augmenting}
\end{figure*}

\subsection{Uniqueness. } Figure \ref{fig1:privacy} shows the average per fold proportion of duplicated data from the generation of $100,000$ samples. Of $100,000$ generated samples, VAE has the highest duplication rate of the original training data ($67\%$), followed by the CTGAN ($34\%$), DPGAN001 ($32\%$) and DPGAN050 ($18\%$), as shown in grey. Although DPGAN050 has the lowest duplication rate of the original training data, its novel synthesised samples are largely duplicated (approx $71$k duplicates shown in light purple), compared to $26$k, $55$k and $31$k for VAE, DPGAN001 and CTGAN, respectively. CTGAN generates the largest number of unique samples at an average of $33,689$, and VAE has the fewest with $6,647$ (dark purple). Both DPGAN models generate a similar number of unique samples ($12,481$ for DPGAN001 and $11,288$ for DPGAN050). 
\subsection{Similarity. } Random subsets from each unique synthetic dataset are sampled, and \textit{Similarity} with the original data is computed (\ref{fig2:similarity}). Batches are sampled to match the size of the training data while preserving the outcome class ratio. Synthesiser VAE has the highest precision ($0.95\pm0.01$) and density ($0.63\pm0.03$). Given that this model has the largest copying rate of the original data, its unique data also lies closest to the original data, preserving fidelity. Synthesiser CTGAN has the lowest fidelity (both precision $0.37\pm0.04$ and density $0.13\pm0.02$) with the original training data; however, it attains the highest recall ($0.67\pm0.02$). CTGAN displays the largest degradation ($0.31\pm0.02$) when measuring diversity using coverage, placing last with DPGAN050 ($0.31\pm0.02$), which suggests that many of the diverse samples generated by CTGAN are considered outliers (they do not match the density of the original distribution). VAE also displays a marked reduction in diversity compared to both DPGAN models when measured with recall ($0.58\pm0.03$) versus coverage ($0.43\pm0.01$). DPGAN001 scores consistently higher than DPGAN050 across all \textit{Similarity} metrics ($0.64\pm0.09$ vs $0.54\pm0.05$ for precision, $0.46\pm0.12$ vs $0.31\pm0.02$ for recall, $0.28\pm0.05$ vs $0.22\pm0.02$ for density and $0.44\pm0.10$ vs $0.31\pm0.12$ for coverage).

\subsection{Utility. } None of the synthetic datasets reaches an AUC-ROC that is higher than the original training data in the experiments with Setting A. VAE has the highest AUC of all synthetic datasets, which is in line with its ranking as the highest across similarity metrics. Model rank across the sum of \textit{Similarity} metrics is VAE, DPGAN001, CTGAN and DPGAN050, which do not follow AUC-ROC scores. Besides, we also compared performance measures of a balanced dataset by both upsampling and downsampling to a balanced dataset augmented with synthetic samples (Setting B). The ordering of AUC-ROC scores is preserved from the synthetic data only classification task (Figure \ref{fig:augauc} versus Figure \ref{fig3:synth_auc}) with a smaller difference from the original baseline performance. In contrast, synthetically augmented data from VAE and CTGAN outperform the baseline models when measured using accuracy (Figure \ref{fig:augacc}). This increased performance is largely driven by recall (true positive rate) for the survived subgroup (Figure \ref{fig:augrec}).

\section{\uppercase{Discussion}}\label{discussion} 

\begin{figure*}
     \begin{subfigure}[b]{0.19\textwidth}
         \centering
         \includegraphics[width=\textwidth]{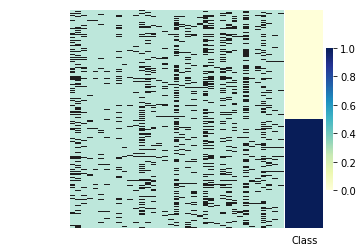}
         \caption{Oversampled}
         \label{fig1:oversampled_heatmap}
     \end{subfigure}
     \hfill
     \begin{subfigure}[b]{0.19\textwidth}
         \centering
         \includegraphics[width=\textwidth]{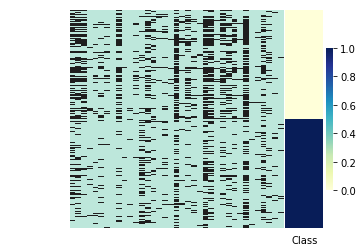} 
         \caption{VAE}
         \label{fig2:vae_heatmap}
     \end{subfigure}
     \begin{subfigure}[b]{0.19\textwidth}
         \centering
         \includegraphics[width=\textwidth]{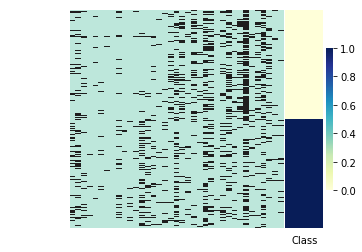}
         \caption{DPGAN050}
         \label{fig3:dpgan050_heatmap}
     \end{subfigure}
     \begin{subfigure}[b]{0.19\textwidth}
         \centering
         \includegraphics[width=\textwidth]{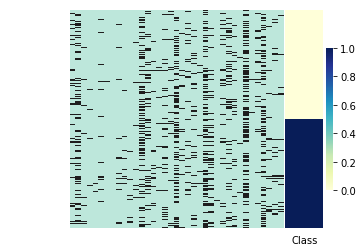}
         \caption{DPGAN001}
         \label{fig3:dpgan001_heatmap}
     \end{subfigure}
     \begin{subfigure}[b]{0.19\textwidth}
         \centering
         \includegraphics[width=\textwidth]{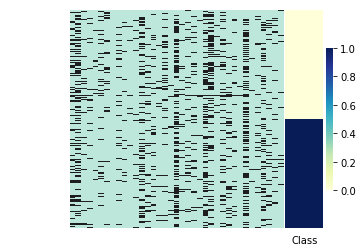}
         \caption{CTGAN}
         \label{fig3:ctgan_heatmap}
     \end{subfigure}
     \hfill
     \begin{subfigure}[b]{0.19\textwidth}
         \centering
         \includegraphics[width=0.9\textwidth]{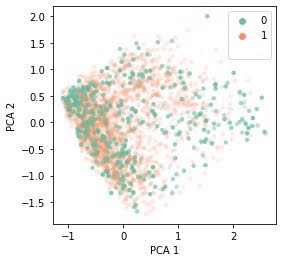}
         \caption{Oversampled}
         \label{fig1:oversampled_pca}
     \end{subfigure}
     \hfill
     \begin{subfigure}[b]{0.19\textwidth}
         \centering
         \includegraphics[width=0.9\textwidth]{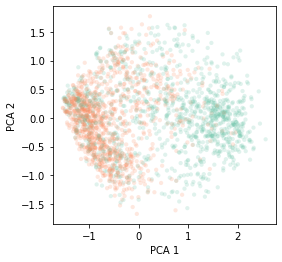} 
         \caption{VAE}
         \label{fig2:vae_pca}
     \end{subfigure}
     \begin{subfigure}[b]{0.19\textwidth}
         \centering
         \includegraphics[width=0.9\textwidth]{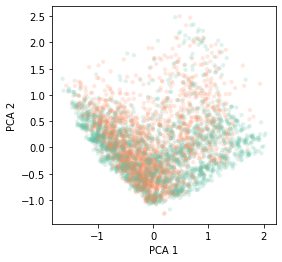}
         \caption{DPGAN050}
         \label{fig3:dpgan050_pca}
     \end{subfigure}
     \begin{subfigure}[b]{0.19\textwidth}
         \centering
         \includegraphics[width=0.9\textwidth]{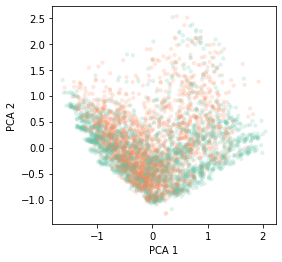}
         \caption{DPGAN001}
         \label{fig3:dpgan001_pca}
     \end{subfigure}
     \begin{subfigure}[b]{0.19\textwidth}
         \centering
         \includegraphics[width=0.9\textwidth]{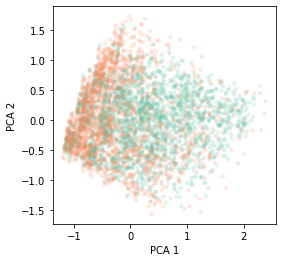}
         \caption{CTGAN}
         \label{fig3:ctgan_pca}
     \end{subfigure}
    \caption{The top row shows the heatmaps of binary matrices of sampled examples, where the columns represent different features and rows represent different samples. Class 0 at the top half of the matrix represents examples sampled from the dataset augmented with synthetic data, and Class 1 at the bottom half represents examples sampled from the original dataset. The balanced datasets serve as input for the classification algorithms as per \nameref{results} section. The bottom row shows 2D PCA embeddings of the same data.}
    \label{fig:sampling_bias}
\end{figure*}

\noindent The increased predictive performance with synthetically augmented data from VAE and CTGAN is driven by the recall, indicating that these models can better identify the positive class with little degradation to the precision. CTGAN has the most significant number of unique samples and has the most remarkable diversity when measured using recall. This diversity of CTGAN lends itself to extracting generalisable features which predict each class. While VAE produces far fewer unique samples, it retains diversity in its samples. To better understand the nature of the signal identified by these models, we have plotted the heatmap of the binary feature matrix for one fold (see Figure \ref{fig:sampling_bias}). In the top row of figures, the lower half of the heatmap comes from the original distribution (marked as class 1). In contrast, the upper half is original data augmented with synthetic data for class balance (marked as class 0). Here we observe that synthesiser VAE has largely exaggerated the signal for features in augmented data (class 0), which has resulted in greater separability in the input space, as shown by the two-dimension PCA reduction. Both DPGAN models display reduced separability, with DPGAN001 reporting a lower accuracy. The heatmaps go some way in explaining this since there does not appear to be a marked difference in the upper and lower halves of the DPGAN001 heatmap. DPGAN050, however, does show some exaggerated features for class 0. Visually, CTGAN appears to have the most similar heatmap as the original data upsampled while obtaining greater separability. This visual heuristic of similarity appears to contrast the \textit{Similarity} metrics reported in the previous section. Since the \textit{Similarity} metrics measure both the positive and negative outcome class in proportions with which they appear in the original training set, the similarity of the larger positive class could have far outweighed that of the smaller negative class.

\section{\uppercase{Conclutions}}
\noindent This work has examined the \textit{Utility}, \textit{Uniqueness} and \textit{Similarity} of the synthetic Cystic Fibrosis patients' EHRs with four SOTA generative models. We observed increased accuracy in predictive performance with both VAE and CTGAN when we augmented the EHRs with synthetic data. \textit{Similarity} metrics appear to explain the \textit{Utility} performance of synthetic data generators. While the amplification of a signal in the synthetic dataset may do poorly to preserve the faithfulness of the original data, it can provide greater separability hence predictive performance. Considering the \textit{Uniqueness} of each synthetic data generator, CTGAN offers both high predictive performance and \textit{Uniqueness} of samples, which is beneficial for considering stricter conditions on privacy. These trade-offs are problem-specific, and conclusions are to be arrived at based on clinical relevance. For example, do exaggerated signals corroborate clinical evidence for co-occurrence of diseases? Given the caveats of our dataset, we have not shown the clinical relevance of individual features, which remains as future work.

\bibliographystyle{apalike}
{\small \bibliography{main}}

\begin{thebibliography}{}

\bibitem[Alaa et~al., 2021]{alaa2021faithful}
Alaa, A.~M., van Breugel, B., Saveliev, E., and van~der Schaar, M. (2021).
\newblock How faithful is your synthetic data? sample-level metrics for
  evaluating and auditing generative models.
\newblock {\em arXiv preprint arXiv:2102.08921}.

\bibitem[Chen et~al., 2021]{chen2021synthetic}
Chen, R.~J., Lu, M.~Y., Chen, T.~Y., Williamson, D.~F., and Mahmood, F. (2021).
\newblock Synthetic data in machine learning for medicine and healthcare.
\newblock {\em Nature Biomedical Engineering}, pages 1--5.

\bibitem[Choi et~al., 2017]{choi2017generating}
Choi, E., Biswal, S., Malin, B., Duke, J., Stewart, W.~F., and Sun, J. (2017).
\newblock In {\em Machine learning for healthcare conference}, pages 286--305.
  PMLR.

\bibitem[Douzas and Bacao, 2018]{douzas2018effective}
Douzas, G. and Bacao, F. (2018).
\newblock Effective data generation for imbalanced learning using conditional
  generative adversarial networks.
\newblock {\em Expert Systems with applications}, 91:464--471.

\bibitem[Dwork et~al., 2014]{dwork2014algorithmic}
Dwork, C., Roth, A., et~al. (2014).
\newblock The algorithmic foundations of differential privacy.
\newblock {\em Found. Trends Theor. Comput. Sci.}, 9(3-4):211--407.

\bibitem[Engelmann and Lessmann, 2020]{engelmann2020conditional}
Engelmann, J. and Lessmann, S. (2020).
\newblock Conditional wasserstein gan-based oversampling of tabular data for
  imbalanced learning.
\newblock {\em arXiv preprint arXiv:2008.09202}.

\bibitem[Fiore et~al., 2019]{fiore2019using}
Fiore, U., De~Santis, A., Perla, F., Zanetti, P., and Palmieri, F. (2019).
\newblock Using generative adversarial networks for improving classification
  effectiveness in credit card fraud detection.
\newblock {\em Information Sciences}, 479:448--455.

\bibitem[Kynk{\"a}{\"a}nniemi et~al., 2019]{kynkaanniemi2019improved}
Kynk{\"a}{\"a}nniemi, T., Karras, T., Laine, S., Lehtinen, J., and Aila, T.
  (2019).
\newblock Improved precision/recall metric for assessing generative models.
\newblock {\em arXiv preprint arXiv:1904.06991}.

\bibitem[Liu et~al., 2019]{liu2019wasserstein}
Liu, Y., Zhou, Y., Liu, X., Dong, F., Wang, C., and Wang, Z. (2019).
\newblock Wasserstein gan-based small-sample augmentation for new-generation
  artificial intelligence: a case study of cancer-staging data in biology.
\newblock {\em Engineering}, 5(1):156--163.

\bibitem[Naeem et~al., 2020]{naeem2020reliable}
Naeem, M.~F., Oh, S.~J., Uh, Y., Choi, Y., and Yoo, J. (2020).
\newblock Reliable fidelity and diversity metrics for generative models.
\newblock In {\em International Conference on Machine Learning}, pages
  7176--7185. PMLR.

\bibitem[Ngwenduna and Mbuvha, 2021]{ngwenduna2021alleviating}
Ngwenduna, K.~S. and Mbuvha, R. (2021).
\newblock Alleviating class imbalance in actuarial applications using
  generative adversarial networks.
\newblock {\em Risks}, 9(3):49.

\bibitem[Sajjadi et~al., 2018]{sajjadi2018assessing}
Sajjadi, M.~S., Bachem, O., Lucic, M., Bousquet, O., and Gelly, S. (2018).
\newblock Assessing generative models via precision and recall.
\newblock {\em arXiv preprint arXiv:1806.00035}.

\bibitem[USCFF, 2020]{USCF}
USCFF (2020).
\newblock Patient registry annual data report.
\newblock \url{https://www.cff.org/media/23476/download}.
\newblock Accessed: 07-12-2021.

\bibitem[Xie et~al., 2018]{xie2018differentially}
Xie, L., Lin, K., Wang, S., Wang, F., and Zhou, J. (2018).
\newblock Differentially private generative adversarial network.
\newblock {\em arXiv preprint arXiv:1802.06739}.

\bibitem[Xu et~al., 2019]{xu2019modeling}
Xu, L., Skoularidou, M., Cuesta-Infante, A., and Veeramachaneni, K. (2019).
\newblock Modeling tabular data using conditional gan.
\newblock {\em arXiv preprint arXiv:1907.00503}.

\bibitem[Xu and Veeramachaneni, 2018]{xu2018synthesizing}
Xu, L. and Veeramachaneni, K. (2018).
\newblock Synthesizing tabular data using generative adversarial networks.
\newblock {\em arXiv preprint arXiv:1811.11264}.

\bibitem[Yoon et~al., 2020]{yoon2020anonymization}
Yoon, J., Drumright, L.~N., and Van Der~Schaar, M. (2020).
\newblock Anonymization through data synthesis using generative adversarial
  networks (ads-gan).
\newblock {\em IEEE journal of biomedical and health informatics},
  24(8):2378--2388.

\end{thebibliography}

\end{document}